\documentclass[runningheads]{llncs}

\usepackage[width=122mm,left=12mm,paperwidth=146mm,height=193mm,top=12mm,paperheight=217mm]{geometry}

\usepackage{graphicx}
\usepackage{comment}
\usepackage{amsmath,amssymb} 
\usepackage{color}
\usepackage{cite}
\usepackage{microtype}
\usepackage[pagebackref=true,breaklinks=true,letterpaper=true,colorlinks,bookmarks=false]{hyperref}
\hypersetup{
    colorlinks = true,
    citecolor  = blue,
    linkcolor  = blue,
    urlcolor   = blue,
}
\usepackage{cleveref}
\Crefformat{figure}{Fig.~#2#1#3}                           
\Crefname{subfigure}{Fig.}{Figs.}
\Crefname{figure}{Fig.}{Figs.}
\usepackage{courier}
\usepackage{booktabs} 
\renewcommand{\vec}[1]{\boldsymbol{#1}}

\usepackage{subfig}
\DeclareRobustCommand\onedot{\futurelet\@let@token\@onedot}
\def\@onedot{\ifx\@let@token.\else.\null\fi\xspace}

\def\ie{\textit{i.e}}

\begin{document}
\pagestyle{headings}
\mainmatter
\def\ECCVSubNumber{2662}  

\title{
    Tensor Low-Rank Reconstruction for \\ Semantic Segmentation
}

\titlerunning{}
%

\author{
    Wanli Chen\inst{1}\and
    Xinge Zhu\inst{1} \and
    Ruoqi Sun\inst{2} \and
    Junjun He\inst{2,3} \and
    Ruiyu Li\inst{4} \and \\
    Xiaoyong Shen\inst{4} \and
    Bei Yu\inst{1}
}

\authorrunning{W.~Chen et al.}
%
\institute{
    The Chinese University of Hong Kong
    \\ \email{\{wlchen,byu\}@cse.cuhk.edu.hk,zx018@ie.cuhk.edu.hk}
    \and
    Shanghai Jiao Tong University
    \\ \email{ruoqisun7@sjtu.edu.cn} 
    \and
    ShenZhen Key Lab of Computer Vision and Pattern Recognition,
    SIAT-SenseTime Joint Lab, Shenzhen Institutes of Advanced Technology, Chinese Academy of Sciences 
    \\ \email{hejunjun@sjtu.edu.cn}
    \and
    SmartMore
    \\ \email{\{ryli,xiaoyong\}@smartmore.com}
}
\maketitle

\begin{abstract}

    Context information plays an indispensable role in the success of semantic segmentation. Recently, non-local self-attention based methods are proved to be effective for context information collection. Since the desired context consists of spatial-wise and channel-wise attentions, 3D representation is an appropriate formulation. However, these non-local methods describe 3D context information based on a 2D similarity matrix, where space compression may lead to channel-wise attention missing.
    An alternative is to model the contextual information directly without compression.
    However, this effort confronts a fundamental difficulty, namely the high-rank property of context information.
    In this paper, we propose a new approach to model the 3D context representations, which not only avoids the space compression but also tackles the high-rank difficulty. 
    Here, inspired by tensor canonical-polyadic decomposition theory (\ie , a high-rank tensor can be expressed as a combination of rank-1 tensors.), 
    we design a low-rank-to-high-rank context reconstruction framework (\ie , RecoNet). 
    Specifically, we first introduce the tensor generation module (TGM), which generates a number of rank-1 tensors to capture fragments of context feature.
    Then we use these rank-1 tensors to recover the high-rank context features through our proposed tensor reconstruction module (TRM).
    Extensive experiments show that our method achieves state-of-the-art on various public datasets.
    Additionally, our proposed method has more than 100 times less computational cost compared with conventional non-local-based methods.
    
\end{abstract}





\section{Introduction}

\label{sec:intro}

Semantic segmentation aims to assign the pixel-wise predictions for the given image, which is a challenging task requiring fine-grained shape, texture and category recognition.
The pioneering work, fully convolutional networks (FCN)~\cite{SEGM-CVPR2015-FCN}, explores the effectiveness of deep convolutional networks in segmentation task. 
Recently, more work achieves great progress from exploring the contextual information
\cite{SEGM-MICCAI2015-UNet,SEGM-TPAMI2017-SegNet,SEGM-CVPR2017-RefineNet,SEGM-arXiv2017-DeepLabv3,SEGM-CVPR2017-PSPNet,SEGM-ECCV2018-DeepLabv3+,sun2019not},
in which non-local based methods are the recent mainstream~\cite{SEGM-ECCV2018-PSANet,SEGM-CVPR2019-CFNet,SEGM-CVPR2019-APCNet}.
\begin{figure}[tb!] 
    \centering 
    \subfloat[]{
        \includegraphics[width=0.718\linewidth]{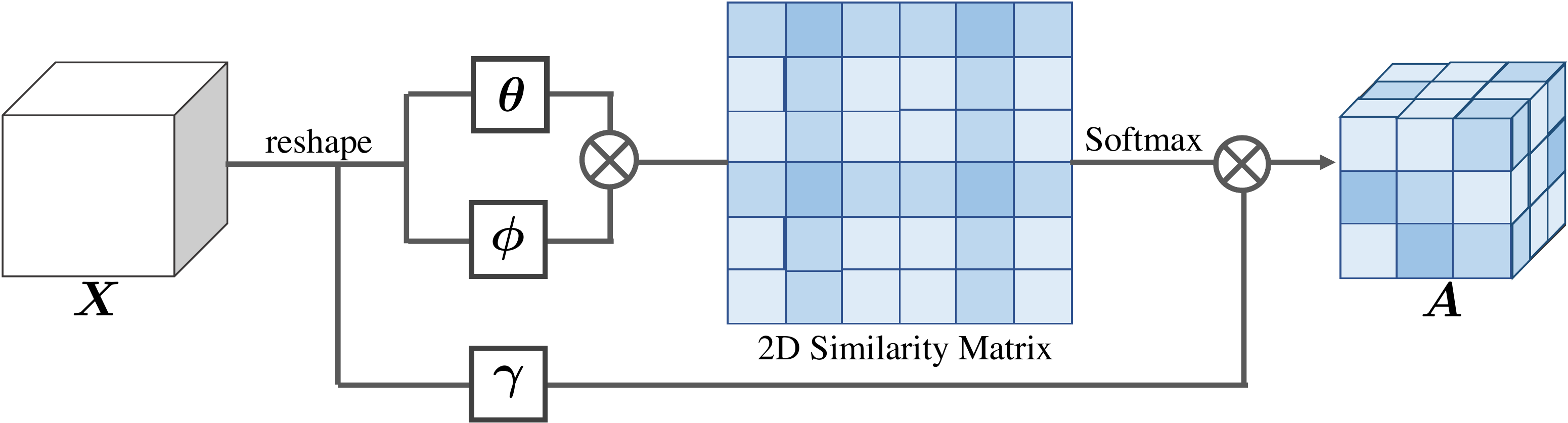}  
        \label{fig:reconstruct-1}
    }\\
    \subfloat[]{
        \includegraphics[width=0.78\linewidth]{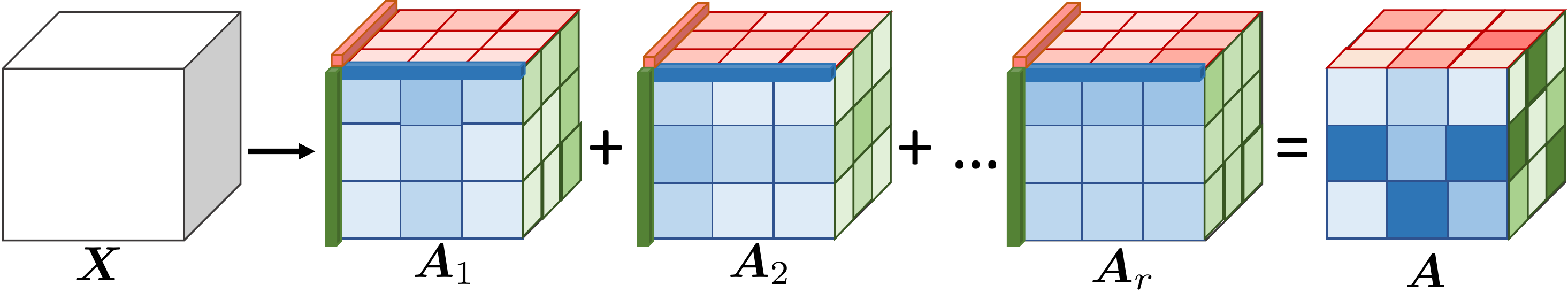}
        \label{fig:reconstruct-2}
    }
    \caption{
        (a) Non-local vs.~(b) our proposed RecoNet, which is based on tensor low-rank reconstruction.
        Note that 2D similarity matrix exists in non-local based methods and our RecoNet is formed with all 3D tensors.
    }
    \label{fig:context_aggregation_model}
\end{figure}
These methods model the context representation by rating the element-wise importance for contextual tensors. 
However, the context features obtained from this line lack of channel-wise attention, which is a key component of context. 
Specifically, for a typical non-local block, the 2D similarity map $\vec{A} \in \mathbb{R}^{HW \times HW}$ is generated by the matrix multiplication of two inputs with dimension of $H \times W \times C$ and $C \times H\times W$, respectively.
It is noted that the channel dimension $C$ is eliminated during the multiplication,
which implies that only the spatial-wise attention is represented while the channel-wise attention is compressed.
Therefore, these non-local based methods could collect fine-grained spatial context features but may sacrifice channel-wise context attention.

An intuitive idea tackling this issue is to construct the context directly instead of using the 2D similarity map.
Unfortunately, this approach confronts a fundamental difficulty because of the high-rank property of context features~\cite{SEGM-CVPR2019-CFNet}.
That is, the context tensor should be high-rank to have enough capacity since contexts vary from image to image and this large diversity cannot be well-represented by very limited parameters. 

Inspired by tensor canonical-polyadic decomposition theory~\cite{TNSR-SIREV2009-Kolda}, \textit{i.e.}, a high-rank tensor can be expressed as a combination of rank-1 tensors, 
we propose a new approach of modeling high-rank contextual information in a progressive manner without channel-wise space compression.
We show the workflow of non-local networks and RecoNet in \Cref{fig:context_aggregation_model}.
The basic idea is to first use a series of low-rank tensors to collect fragments of context features and then build them up to reconstruct fine-grained context features.
Specifically, our proposed framework consists of two key components, rank-1 tensor generation module (TGM) and high-rank tensor reconstruction module (TRM).
Here, TGM aims to generate the rank-1 tensors in channel, height and width directions, which explore the context features in different views with low-rank constraints.
TRM adopts tensor canonical-polyadic (CP) reconstruction to reconstruct the high-rank attention map,
in which the co-occurrence contextual information is mined based on the rank-1 tensors from different views.
The cooperation of these two components leads to the effective and efficient high-rank context modeling.

We tested our method on five public datasets.
On these experiments, the proposed method consistently achieves the state-of-the-art, especially for PASCAL-VOC12~\cite{CV-IJCV2010-VOC}, RecoNet reaches the \textbf{top-1} performance.
Furthermore, by incorporating the simple and clean low-rank features, our whole model has less computation consumption (more than \textbf{100} times lower than non-local) compared to other non-local based context modeling methods.


The contributions of this work mainly lie in three aspects: 
\begin{itemize}
    \item Our studies reveal a new path to the context modeling, namely, context reconstruction from low-rank to high-rank in a progressive way.
 	\item We develop a new semantic segmentation framework RecoNet, which explores the contextual information through tensor CP reconstruction.
        It not only keeps both spatial-wise and channel-wise attentions, but also deals with high-rank difficulty.
 	\item We conduct extensive experiments to compare the proposed methods with others on various public datasets, where it yields notable performance gains.
        Furthermore, RecoNet also has less computation cost, \ie, more than \textbf{100} times smaller than non-local based methods.
\end{itemize}

\section{Related Work}

\noindent\textbf{Tensor Low-rank Representation.}~~~
According to tensor decomposition theory~\cite{TNSR-SIREV2009-Kolda}, a tensor can be represented by the linear combination of series of low-rank tensors.
The reconstruction results of these low-rank tensors are the principal components of original tensor.
Therefore, tensor low-rank representation is widely used in computer vision task such as convolution speed-up~\cite{SPEED-ICLR2015-Lebedev} and model compression~\cite{SPEED-CVPR2017-Yu}.
There are two tensor decomposition methods: Tuker decomposition and CP decompostion~\cite{TNSR-SIREV2009-Kolda}.
For the Tuker decomposition, the tensor is decomposed into a set of matrices and one core tensor.
If the core tensor is diagonal, then Tuker decomposition degrades to CP decomposition.
For the CP decomposition, the tensor is represented by a set of rank-1 tensors (vectors).
In this paper, we apply this theory for \emph{reconstruction}, namely reconstructing high-rank contextual tensor from a set of rank-1 context fragments.

\noindent\textbf{Self-Attention in Computer Vision.}~~~
Self attention is firstly proposed in natural language processing (NLP)~\cite{ATTN-NIPS2017-Vaswani,ATTN-NIPS2015-Chorowski,ATTN-NAACL2016-Yang,ATTN-ACL2017-Cui}.
It serves as a global encoding method that can merge long distance features. This property is also important to computer vision tasks.
Hu \textit{et al}.~propose SE-Net~\cite{IMGC-CVPR2018-SENet}, exploiting channel information for better image classification through channel wise attention.
Woo \textit{et al}.~propose CBAM~\cite{ATTN-ECCV2018-CBAM} that combines channel-wise attention and spatial-wise attention to capture rich feature in CNN.
Wang \textit{et al}.~propose non-local neural network~\cite{ATTN-CVPR2018-Wang}.
It catches long-range dependencies of a featuremap, which breaks the receptive field limitation of convolution kernel.

\noindent\textbf{Context Aggregation in Semantic Segmentation.}~~~
Context information is so important for semantic segmentation and many researchers pay their attention to explore the context aggregation.
The initial context harvesting method is to increase receptive fields such as FCN~\cite{SEGM-CVPR2015-FCN}, which merges feature of different scales.
Then feature pyramid methods~\cite{SEGM-arXiv2017-DeepLabv3,SEGM-CVPR2017-PSPNet,SEGM-ECCV2018-DeepLabv3+} are proposed for better context collection.
Although feature pyramid collects rich context information, the contexts are not gathered adaptively.
In other words, the importance of each element in contextual tensor is not discriminated.
Self-attention-based methods are thus proposed to overcome this problem,
such as EncNet~\cite{SEGM-CVPR2018-EncNet}, PSANet~\cite{SEGM-ECCV2018-PSANet}, APCNet~\cite{SEGM-CVPR2019-APCNet}, and CFNet~\cite{SEGM-CVPR2019-CFNet}.
Researchers also propose some efficient self-attention methods such as EMANet~\cite{SEGM-ICCV2019-EMANet}, CCNet~\cite{SEGM-ICCV2019-CCNet}, $A^2$Net~\cite{ATTN-NIPS2018-A2-Nets},
which have lower computation consumption and GPU memory occupation.
However, most of these methods suffer from channel-wise space compression due to the 2D similarity map.
Compared to these works, our method differs essentially in that it uses the 3D low-rank tensor reconstruction to catch long-range dependencies without sacrificing channel-wise attention.

\section{Methodology}

\subsection{Overview}
The semantic information prediction from an image is closely related to the context information.
Due to the large varieties of context, a high-rank tensor is required for the context feature representation. 
However, under this constraint, modeling the context features directly means a huge cost.  
Inspired by the CP decomposition theory, although the context prediction is a high-rank problem,
we can separate it into a series of low-rank problems and these low-rank problems are easier to deal with.
Specifically, we do not predict context feature directly, instead, we generate its fragments.
Then we build up a complete context feature using these fragments.
The low-rank to high-rank reconstruction strategy not only maintains 3D representation (for both channel-wise and spatial-wise), but also tackles with the high-rank difficulty. 

\begin{figure*}[tb!] 
    \centering 
    \includegraphics[width=1.0\linewidth]{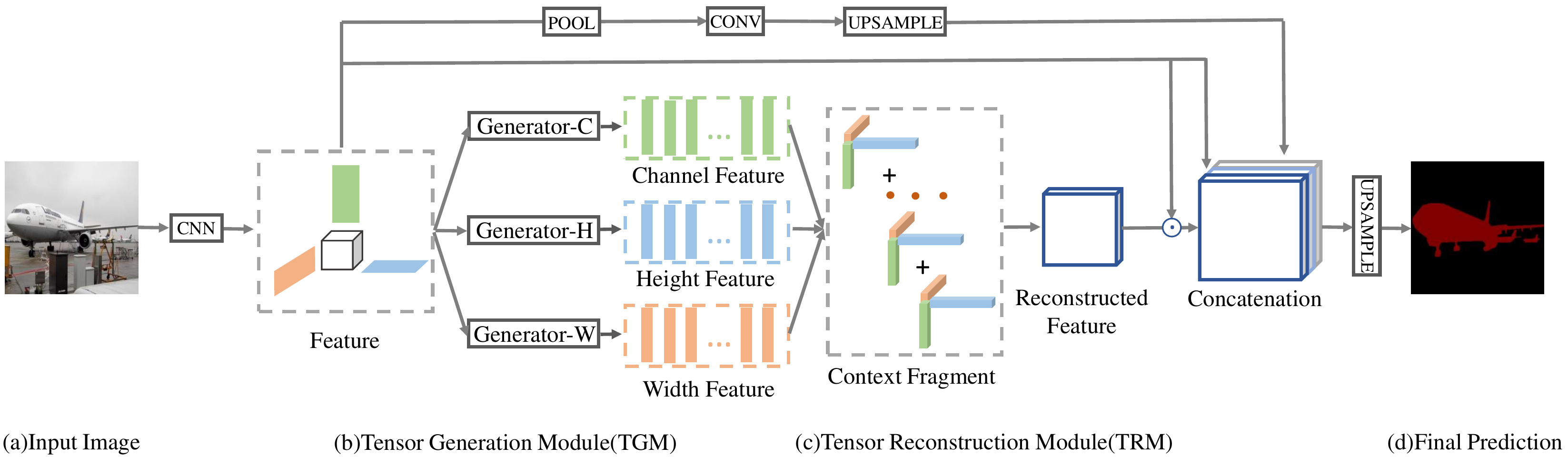} 
    \caption{The pipeline of our framework. Two major components are involved, \ie, Tensor Generation Module (TGM) and Tensor Reconstruction Module (TRM). TGM performs the low-rank tensor generation while TRM achieves the high-rank tensor reconstruction via CP construction theory.} 
    \label{fig:main_frame}
\end{figure*}

The pipeline of our model is shown in \Cref{fig:main_frame}, which consists of low-rank tensor generation module (TGM), high-rank tensor reconstruction module (TRM), and global pooling module (GPM) to harvest global context in both spatial and channel dimensions.
We upsample the model output using bilinear interpolation before semantic label prediction. 


In our implementation, multiple low-rank perceptrons are used to deal with the high-rank problem, by which we learn parts of context information (\ie, context fragments).
We then build the high-rank tensor via tensor reconstruction theory \cite{TNSR-SIREV2009-Kolda}.

\noindent \textbf{Formulation:}~~~ Assuming we have 3$r$ vectors in C/H/W directions $\vec{v}_{ci} \in \mathbb{R}^{C \times 1 \times 1}$, $\vec{v}_{hi} \in \mathbb{R}^{1 \times H \times 1}$ and $\vec{v}_{wi} \in \mathbb{R}^{1 \times 1 \times W}$, where $i \in r$ and $r$ is the tensor rank. These vectors are the CP decomposed fragments of  $\vec{A} \in \mathbb{R}^{C \times H \times W}$ , then tensor CP rank-$r$ reconstruction is defined as:
\begin{align} 
\label{equ:TensorRecon}
\vec{A} = \sum_{i=1}^{r} \lambda_i \vec{v}_{ci} \otimes \vec{v}_{hi} \otimes \vec{v}_{wi},
\end{align}
where $\lambda_i$ is a scaling factor.

\subsection{Tensor Generation Module}

In this section, we first provide some basic definitions and then show how to derive the low-rank tensors from the proposed module.

\noindent \textbf{Context Fragments.}~~~
We define context fragments as the outputs of the tensor generation module,
which indicates some rank-1 vectors $\vec{v}_{ci}$, $\vec{v}_{hi}$ and $\vec{v}_{wi}$ (as defined in previous part) in the channel, the height and the width directions.
Every context fragment contains a part of context information.

\noindent \textbf{Feature Generator.}~~~
We define three feature generators:
Channel Generator, Height Generator and Width Generator.
Each generator is composed of Pool-Conv-Sigmoid sequence.
Global pooling is widely used in previous works \cite{SEGM-arXiv2015-ParseNet, SEGM-CVPR2017-PSPNet} as the global context harvesting method.
Similarly, here we use global average pooling in feature generators, obtaining the global context representation in C/H/W directions. 

\begin{figure}[t] 
    \centering 
    \includegraphics[width=0.68\linewidth]{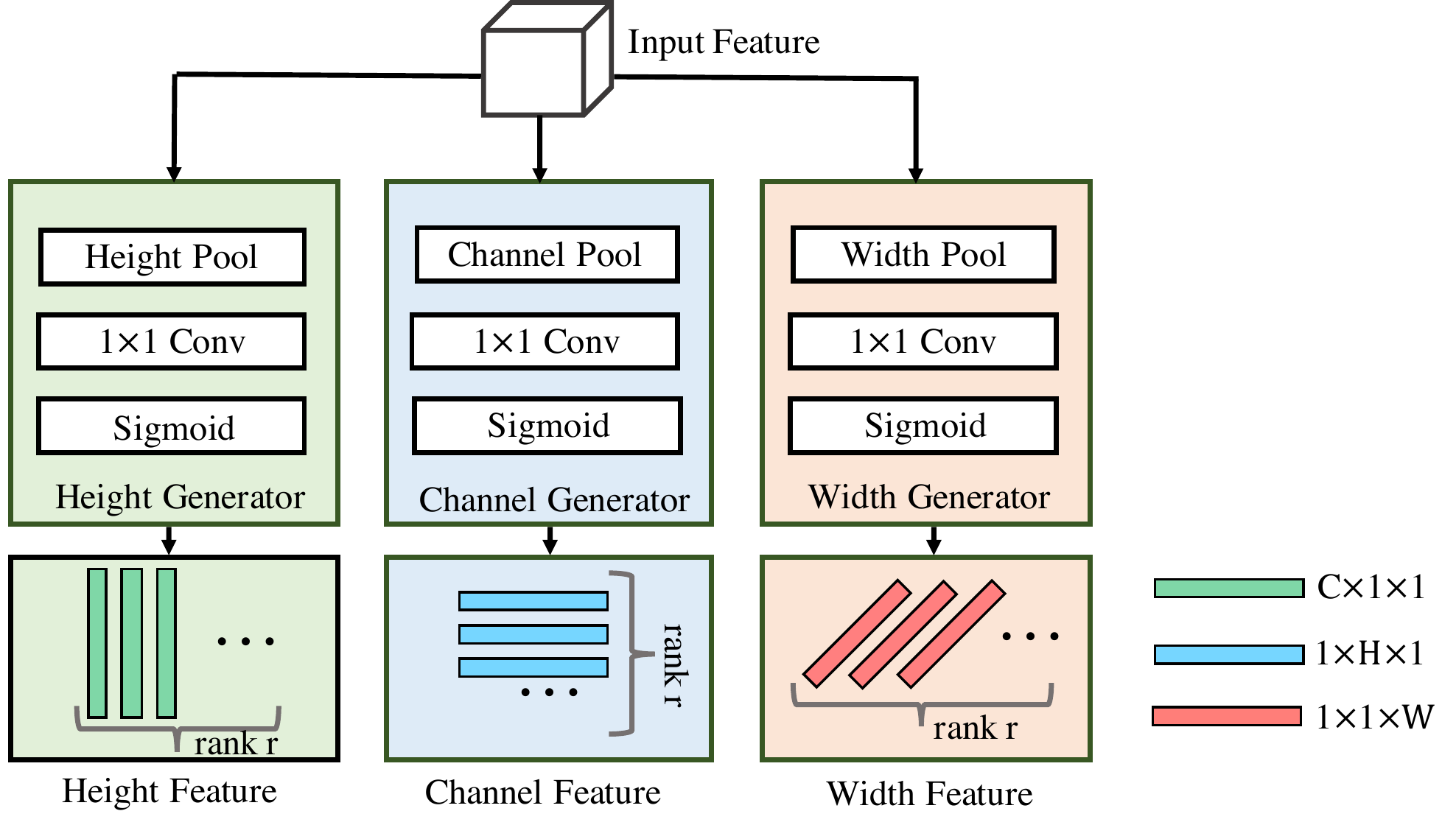} 
    \caption{Tensor Generation Module. Channel Pool, Height Pool and Width Pool are all global average pooling.}  
    \label{fig:TGM}
\end{figure}

\noindent \textbf{Context Fragments Generation.}~~~
In order to learn fragments of context information across the three directions, we apply channel, height and width generator on the top of input feature.
We repeat this process $r$ times obtaining 3$r$ learnable vectors $\vec{v}_{ci} \in \mathbb{R}^{C \times 1 \times 1}$, $\vec{v}_{hi} \in \mathbb{R}^{1 \times H \times 1}$ and $\vec{v}_{wi} \in \mathbb{R}^{1 \times 1 \times W}$, where $i \in r$. All vectors are generated using independent convolution kernels. Each of them learns a part of context information and outputs as context fragment. The TGM is shown in \Cref{fig:TGM}.

\noindent \textbf{Non-linearity in TGM.}~~~
Recalling that TGM generates 3$r$ rank-1 tensors and these tensors are activated by $\mathrm{Sigmoid}$ function, which re-scales the values in context fragments to [0, 1]. We add the non-linearity for two reasons. Firstly, each re-scaled element can be regarded as the weight of a certain kind of context feature, which satisfy the definition of attention. Secondly, all the context fragments shall not be linear dependent so that each of them can represent different information.

\subsection{Tensor Reconstruction Module}


In this part, we introduce the context feature reconstruction and aggregation procedure. The entire reconstruction process is clean and simple, which is based on \Cref{equ:TensorRecon}.
For a better interpretation, we first introduce the context aggregation process.

\noindent \textbf{Context Aggregation.}~~~
Different from previous works that only collect spatial or channel attention \cite{SEGM-ECCV2018-PSANet, SEGM-CVPR2018-EncNet}, we collect attention distribution in both directions simultaneously.
The goal of TRM is to obtain the 3D attention map $\vec{A} \in \mathbb{R}^{C \times H \times W}$ which keeps response in both spatial and channel attention. 
After that, context feature is obtained by element-wise product. Specifically, given an input feature $\vec{X}= \lbrace x_1, x_2, \dots, x_{CHW} \rbrace$ and a context attention map $\vec{A}= \lbrace a_1, a_2, \dots, a_{CHW} \rbrace$, the fine-grained context feature $\vec{Y}= \lbrace y_1, y_2, \dots, y_{CHW} \rbrace$ is then given by:
\begin{align} 
\label{equ:aggregation}
\vec{Y} = \vec{A} \cdot \vec{X} \iff y_i = a_i \cdot x_i, i \in CHW.
\end{align}
In this process, every $a_{i} \in \vec{A}$ represents the extent that $x_{i} \in \vec{X}$ be activated. 

\begin{figure*}[htbp] 
    \centering 
    \includegraphics[width=0.78\linewidth]{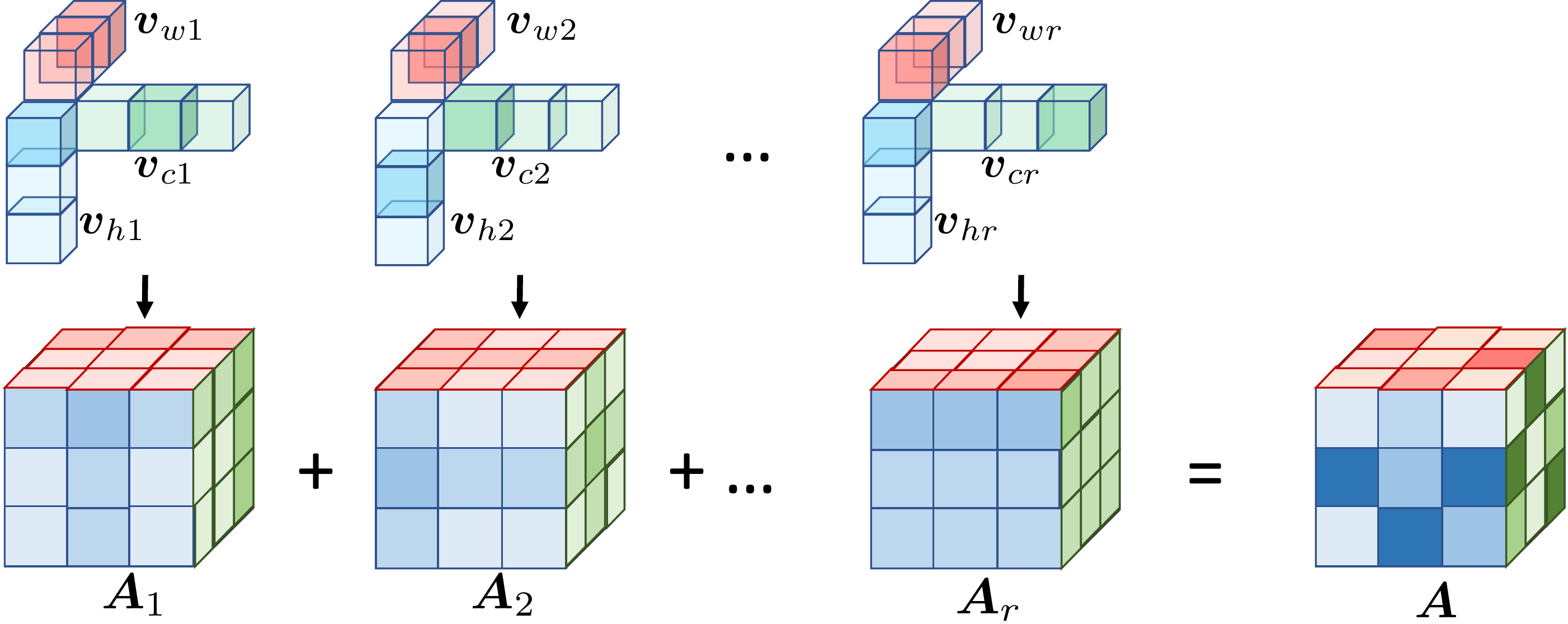} 
    \caption{Tensor Reconstruction Module (TRM). The pipeline of TRM consists of two main steps, \ie, sub-attention map generation and global context reconstruction. The processing from top to bottom (see $\downarrow$) indicates the sub-attention map generation from three dimensions (channel / height / width). The processing from left to right
    (see $\vec{A}_1+\vec{A}_2 + \dots + \vec{A}_r = \vec{A}$ ) denotes the global context reconstruction from low-rank to high-rank. } 
    \label{fig:TRM}
\end{figure*}

\noindent \textbf{Low-rank Reconstruction.}~~~
The tensor reconstruction module (TRM) tackles the high-rank property of context feature.
The full workflow of TRM is shown in~\Cref{fig:TRM}, which consists of two steps, \ie, sub-attention map aggregation and global context feature reconstruction. 
Firstly, three context fragments $\vec{v}_{c1} \in \mathbb{R}^{C \times 1 \times 1}$,
$\vec{v}_{h1} \in \mathbb{R}^{1 \times H \times 1}$ and $\vec{v}_{w1} \in \mathbb{R}^{1 \times 1 \times W}$ are synthesized into a rank-1 sub-attention map $\vec{A}_1$.
This sub-attention map represents a part of 3D context feature, and we will show the visualization of some $\vec{A}_i, i \in[1,r]$ in experimental result part.
Then, other context fragments are reconstructed following the same process. After that we aggregate these sub-attention maps using weighted mean:
\begin{align} 
    \label{equ:recon}
    \vec{A} = \sum_{i=1}^{r} {\lambda}_i \vec{A}_i.
\end{align}
Here ${\lambda_i \in (0, 1)}$ is a learnable normalize factor.
Although each sub-attention map represents low-rank context information, the combination of them becomes a high-rank tensor. The fine-grained context features in both spatial and channel dimensions are obtained after \Cref{equ:recon} and \Cref{equ:aggregation}.

\subsection{Global Pooling Module}
Global pooling module (GPM) is commonly used in previous work \cite{SEGM-CVPR2017-PSPNet,SEGM-CVPR2019-CFNet}. It is composed of a global average pooling operation followed with a 1 $\times$ 1 convolution. It harvests global context in both spatial and channel dimensions. In our proposed model, we apply GPM for the further boost of network performance. 

\subsection{Network Details}

We use ResNet \cite{IMGC-CVPR2016-ResNet} as our backbone and apply dilation strategy to the output of Res-4 and Res-5 of it.
Then, the output stride of our proposed network is 8. The output feature of Res-5 block is marked as $X$.
TGM+TRM and GPM are then added on the top of $X$.
Following previous works \cite{SEGM-CVPR2018-EncNet, SEGM-CVPR2017-PSPNet}, we also use auxiliary loss after Res-4 block.
We set the weight $\alpha$ to 0.2.
The total loss $\mathcal{L}$ is formulated as follows:
\begin{align}
     \mathcal{L} =  \mathcal{L}_{main} + \alpha \mathcal{L}_{aux}.
\end{align}
Finally, we concatenate $X$ with the context featuremap generated by TGM+TRM and the global context generated by GPM to make the final prediction.

\subsection{Relation to Previous Approaches}
Compared with non-local and its variants that explore the pairwise relationship between pixels, the proposed method is essentially unary attention. Unary attention has been widely used in image classification such as SENet \cite{IMGC-CVPR2018-SENet} and CBAM \cite{ATTN-ECCV2018-CBAM}. It is also broadly adopted in semantic segmentation such as DFN \cite{SEGM-CVPR2018-DFN} and EncNet \cite{SEGM-CVPR2018-EncNet}. 
Apparently, SENet is the simplest formation of RecoNet. The 3D attention map of SENet $\vec{A}_{SE} \in \mathbb{R}^{C \times H \times W}$ is as Formula \eqref{equ:SENet}:
\begin{equation}
    \begin{aligned} 
        \label{equ:SENet}
        \vec{A}_{SE} &= \vec{v}_{c} \otimes \vec{v}_{h} \otimes \vec{v}_{w}, \\ 
        \vec{v}_{h} &= \vec{e}, \\ 
        \vec{v}_{w} &= \vec{e}^\top, \\ 
        \vec{e}&=\lbrace 1, 1, 1, \dots 1 \rbrace.
    \end{aligned}
\end{equation}
RecoNet degenerates to SENet by setting tensor rank $r = 1$.
Meanwhile, $\vec{v}_{h} = \vec{e}$ and $\vec{v}_{w} = \vec{e}^\top$.
From Formula \eqref{equ:SENet}, it is observed that the weights in H and W directions are the same, which implies that SENet only harvests channel attention while sets the same weights in spatial domain.
EncNet \cite{SEGM-CVPR2018-EncNet} is the updated version of SENet, which also uses the same spatial weights.
Different spatial weights are introduced in CBAM, which extends Formula \eqref{equ:SENet} to \Cref{equ:CBAM}.
\begin{align} 
    \label{equ:CBAM}
\vec{A}_{CBAM} = \vec{v}_{c} \otimes \vec{v}_{h, w}, ~~~& \vec{v}_{h,w} \in \mathbb{R}^{1 \times H \times W}.
\end{align}
Here $\vec{A}_{CBAM} \in \mathbb{R}^{C \times H \times W}$ is the 3D attention map of CBAM. The spatial attention is considered in CBAM. 
However, single rank-1 tensor $\vec{A}_{CBAM}$ can not represent complicated context information. Considering an extreme case, the spatial attention is CP-decomposed into 2 rank-1 tensors $\vec{v}_{h}\in \mathbb{R}^{1 \times H \times 1}$ and $\vec{v}_{w}\in \mathbb{R}^{1 \times 1 \times W}$. Then, $\vec{A}_{CBAM}$ becomes a sub-attention map of RecoNet.

Simple but effective is the advantage of unary attentions, but they are also criticized for not being able to represent complicated features or for being able to represent features only in one direction (spatial/channel). 
RecoNet not only takes the advantage of simplicity and effectiveness from unary attention, but also delivers comprehensive feature representations from multi-view (\ie, spatial and channel dimension).

\section{Experiments}
Many experiments are carried out in this section. We use five datasets: PASCAL-VOC12, PASCAL-Context, COCO-Stuff, ADE20K and SIFT-FLOW to test the performance of RecoNet.

\subsection{Implementation Details}
RecoNet is implemented using Pytorch \cite{DL-NIPSW2017-PyTorch}.
Following previous works \cite{SEGM-CVPR2019-DANet,SEGM-CVPR2018-EncNet}, synchronized batch normalization is applied.
The learning rate scheduler is $lr=base\_lr \times (1-\cfrac{iter}{total \_ iters}) ^{power}$.
We set $base \_lr$ to 0.001 for PASCAL-VOC12, PASCAL-Context and COCO-Stuff datasets.
The $base \_lr$ for ADE20K and SIFT-FLOW is 0.01 and 0.0025.
Here we set $power$ to 0.9. SGD optimizer is applied with weight decay 0.0001 and momentum 0.9.
We train ADE20K and COCO-Stuff for 120 epochs and 180 epochs respectively. For other datasets, we train 80 epochs.
The batch size we set for all datasets is 16 and all input images are randomly cropped into $512 \times 512$ before putting into neural network.
The data augmentation method we use is the same with previous works \cite{SEGM-CVPR2017-PSPNet,SEGM-CVPR2018-EncNet}.
Specifically, we randomly flip and scale the input images (0.5 to 2).

We use multi-scale and flip evaluation with input scales [0.75, 1, 1.25, 1.5, 1.75, 2.0] times of original scale. The evaluation metrics we use is mean Intersection-over-Union (mIoU).

\subsection{Results on Different Datasets}
\paragraph{\textbf{PASCAL-VOC12}.}
We first test RecoNet using PASCAL-VOC12 \cite{CV-IJCV2010-VOC} dataset, a golden benchmark of semantic segmentation, which includes $20$ object categories and one background class.
The dataset contains $10582$, $1449$, $1456$ images for training, validation and testing.
Our training set contains images from PASCAL augmentation dataset. 
The results are shown in \Cref{tab:VOC_test}. RecoNet reaches $85.6\%$ mIoU, surpassing current best algorithm using ResNet-101 by $1.2\%$, which is a large margin.

\begin{table}[tb!]
    \caption{Results on PASCAL-VOC12 w/o COCO-pretrained model}
    \label{tab:VOC_test}
    \centering
    \resizebox{.98\linewidth}{!}{
        \begin{tabular}{cccccccc}
            \toprule
            &~~FCN\cite{SEGM-CVPR2015-FCN} &~~PSPNet\cite{SEGM-CVPR2017-PSPNet} &~~EncNet\cite{SEGM-CVPR2018-EncNet} &~~APCNet\cite{SEGM-CVPR2019-APCNet} &~~CFNet\cite{SEGM-CVPR2019-CFNet} &~~DMNet\cite{SEGM-ICCV2019-DMNet} &~~RecoNet \\ \midrule
            aero          &76.8          &91.8          &94.1          &95.8          &95.7          &\textbf{96.1} &93.7          \\
            bike          &34.2          &71.9          &69.2          &75.8          &71.9          &\textbf{77.3} &66.3          \\ 
            bird          &68.9          &94.7          &\textbf{96.3} &84.5          &95.0          &94.1          &95.6          \\
            boat          &49.4          &71.2          &\textbf{76.7} &76.0          &76.3          &72.8          &72.8          \\          
            bottle        &60.3          &75.8          &86.2          &80.6          &82.8          &78.1          &\textbf{87.4} \\
            bus           &75.3          &95.2          &96.3          &96.9          &94.8          &\textbf{97.1} &94.5          \\
            car           &74.7          &89.9          &90.7          &90.0          &90.0          &\textbf{92.7} &92.6          \\
            cat           &77.6          &95.9          &94.2          &96.0          &95.9          &96.4          &\textbf{96.5} \\
            chair         &21.4          &39.3          &38.8          &42.0          &37.1          &39.8          &\textbf{48.4} \\
            cow           &62.5          &90.7          &90.7          &93.7          &92.6          &91.4          &\textbf{94.5} \\
            table         &46.8          &71.7          &73.3          &75.4          &73.0          &75.5          &\textbf{76.6} \\
            dog           &71.8          &90.5          &90.0          &91.6          &93.4          &92.7          &\textbf{94.4} \\
            horse         &63.9          &94.5          &92.5          &95.0          &94.6          &95.8          &\textbf{95.9} \\
            mbike         &76.5          &88.8          &88.8          &90.5          &89.6          &91.0          &\textbf{93.8} \\
            person        &73.9          &89.6          &87.9          &89.3          &88.4          &90.3          &\textbf{90.4} \\
            plant         &45.2          &72.8          &68.7          &75.8          &74.9          &76.6          &\textbf{78.1} \\
            sheep         &72.4          &89.6          &92.6          &92.8          &\textbf{95.2} &94.1          &93.6          \\
            sofa          &37.4          &\textbf{64}   &59.0          &61.9          &63.2          &62.1          &63.4          \\
            train         &70.9          &85.1          &86.4          &88.9          &\textbf{89.7} &85.5          &88.6          \\
            tv            &55.1          &76.3          &73.4          &79.6          &78.2          &77.6          &\textbf{83.1} \\ \midrule
            mIoU          &62.2          &82.6          &82.9          &84.2          &84.2          &84.4          &\textbf{85.6} \\
            \bottomrule
        \end{tabular}
    }
\end{table}

\begin{table}[tb!]
    \begin{minipage}{.48\linewidth}

        \caption{Results on PASCAL-VOC w.~COCO-pretrained model}
        \label{tab:VOC_COCO}
        \centering
        \resizebox{4.8cm}{!}{\renewcommand{\arraystretch}{1.0}{
            \begin{tabular}{lc|c}  
                \toprule
                Method                                     &Backbone     & mIoU \\  \hline \hline
                CRF-RNN \cite{SEGM-ICCV2015-Zheng}         &             & 74.7 \\
                DPN \cite{SEGM-ICCV2015-DPN}               &             & 77.5 \\
                Piecewise \cite{SEGM-CVPR2016-Piecewise}   &             & 78.0 \\
                ResNet38 \cite{SEGM-JPR2019-Wu}            &             & 84.9 \\
                PSPNet \cite{SEGM-CVPR2017-PSPNet}         &ResNet-101   & 85.4 \\
                DeepLabv3 \cite{SEGM-arXiv2017-DeepLabv3}  &ResNet-101   & 85.7 \\
                EncNet \cite{SEGM-CVPR2018-EncNet}         &ResNet-101   & 85.9 \\
                DFN \cite{SEGM-CVPR2018-DFN}               &ResNet-101   & 86.2 \\ 
                CFNet \cite{SEGM-CVPR2019-CFNet}           &ResNet-101   & 87.2 \\
                EMANet \cite{SEGM-ICCV2019-EMANet}         &ResNet-101   & 87.7 \\ 
                DeeplabV3+ \cite{SEGM-ECCV2018-DeepLabv3+} &Xception     & 87.8 \\
                DeeplabV3+ \cite{SEGM-ECCV2018-DeepLabv3+} &Xception+JFT & 89.0 \\ \hline \hline
                \textbf{RecoNet}                           &ResNet-101   &$\textbf{88.5}$ \\
                \textbf{RecoNet}                           &ResNet-152   &$\textbf{89.0}$ \\
                \bottomrule
            \end{tabular}
        }}

        \caption{Results on PASCAL-Context test set with background (60 classes)}
        \label{tab:PContext}
        \centering
        \resizebox{4.8cm}{!}{\renewcommand{\arraystretch}{1.0}{
            \begin{tabular}{lc|c}  
                \toprule
                Method &Backbone & mIoU \\  \hline \hline
                FCN-8s \cite{SEGM-CVPR2015-FCN}                &            & 37.8 \\
                ParseNet \cite{SEGM-arXiv2015-ParseNet}        &            & 40.4 \\
                Piecewise \cite{SEGM-CVPR2016-Piecewise}       &            & 43.3 \\
                VeryDeep \cite{SEGM-arXiv2016-Wu}              &            & 44.5 \\
                DeepLab-v2 \cite{SEGM-TPAMI2018-Deeplab}       &ResNet-101  & 45.7 \\
                RefineNet \cite{SEGM-CVPR2017-RefineNet}       &ResNet-152  & 47.3 \\
                PSPNet \cite{SEGM-CVPR2017-PSPNet}             &ResNet-101  & 47.8 \\
                MSCI \cite{SEGM-ECCV2018-MSCI}                 &ResNet-152  & 50.3 \\
                Ding \textit{et al}.~\cite{SEGM-CVPR2018-Ding} &ResNet-101  & 51.6 \\
                EncNet \cite{SEGM-CVPR2018-EncNet}             &ResNet-101  & 51.7 \\
                DANet \cite{SEGM-CVPR2019-DANet}               &ResNet-101  & 52.6 \\
                SVCNet \cite{SEGM-CVPR2019-SVCNet}             &ResNet-101  & 53.2 \\
                CFNet \cite{SEGM-CVPR2019-CFNet}               &ResNet-101  & 54.0 \\
                DMNet \cite{SEGM-ICCV2019-DMNet}               &ResNet-101  & 54.4 \\ \hline \hline
                \textbf{RecoNet}                               &ResNet-101  & \textbf{54.8} \\
                \bottomrule
            \end{tabular}
        }}

    \end{minipage}
    \begin{minipage}{.48\linewidth}

        \caption{Results on COCO-Stuff test set (171 classes)}
        \label{tab:COCO-Stuff}
        \centering
        \resizebox{4.8cm}{!}{\renewcommand{\arraystretch}{1.0}{
            \begin{tabular}{lc|c}  
                \toprule
                Method                                         &Backbone   & mIoU \\  \hline \hline
                FCN-8s \cite{SEGM-CVPR2015-FCN}                &           & 22.7 \\
                DeepLab-v2 \cite{SEGM-TPAMI2018-Deeplab}       &ResNet-101 & 26.9 \\
                RefineNet \cite{SEGM-CVPR2017-RefineNet}       &ResNet-101 & 33.6 \\
                Ding \textit{et al}.~\cite{SEGM-CVPR2018-Ding} &ResNet-101 & 35.7 \\
                SVCNet \cite{SEGM-CVPR2019-SVCNet}             &ResNet-101 & 39.6 \\ 
                DANet \cite{SEGM-CVPR2019-DANet}               &ResNet-101 & 39.7 \\ 
                EMANet \cite{SEGM-ICCV2019-EMANet}             &ResNet-101 & 39.9 \\ \hline \hline
                \textbf{RecoNet}                               &ResNet-101 &\textbf{41.5} \\
                \bottomrule
            \end{tabular}
        }}

        \caption{Results on SIFT-Flow test set}
        \label{tab:SIFT-Flow}
        \centering
        \resizebox{4.8cm}{!}{\renewcommand{\arraystretch}{1.0}{
            \begin{tabular}{l|c|c}
                \toprule
                Method                                         &pixel acc.  & mIoU \\  \hline \hline
                Sharma et al.~\cite{SEGM-NIPS2014-Sharma}      & 79.6       & - \\
                Yang et al.~\cite{SEGM-CVPR2014-Yang}          & 79.8       & - \\
                FCN-8s \cite{SEGM-CVPR2015-FCN}                & 85.9       & 41.2 \\
                DAG-RNN+CRF \cite{SEGM-TPAMI2018-Shuai}        & 87.8       & 44.8 \\
                Piecewise \cite{SEGM-CVPR2016-Piecewise}       & 88.1       & 44.9 \\
                SVCNet \cite{SEGM-CVPR2019-SVCNet}             & 89.1       & 46.3 \\ \hline \hline
                \textbf{RecoNet}                               & \textbf{89.6}       &\textbf{46.8} \\
                \bottomrule
            \end{tabular}
        }}

        \caption{Results on ADE20K $val$ set}
        \label{tab:ADE20K}
        \centering
        \resizebox{4.8cm}{!}{\renewcommand{\arraystretch}{1.0}{
            \begin{tabular}{lc|c}  
                \toprule
                Method                                   & Backbone    & mIoU  \\  \hline \hline
                RefineNet \cite{SEGM-CVPR2017-RefineNet} & ResNet-152  & 40.70 \\
                PSPNet \cite{SEGM-CVPR2017-PSPNet}       & ResNet-101  & 43.29 \\
                DSSPN \cite{SEGM-CVPR2018-DSSPN}         & ResNet-101  & 43.68 \\
                SAC \cite{SEGM-ICCV2017-SAC}             & ResNet-101  & 44.30 \\
                EncNet \cite{SEGM-CVPR2018-EncNet}       & ResNet-101  & 44.65 \\
                CFNet \cite{SEGM-CVPR2019-CFNet}         & ResNet-50   & 42.87 \\
                CFNet \cite{SEGM-CVPR2019-CFNet}         & ResNet-101  & 44.89 \\ 
                CCNet \cite{SEGM-ICCV2019-CCNet}             & ResNet-101  & 45.22 \\
                \hline \hline
                \textbf{RecoNet}                         &ResNet-50    & \textbf{43.40} \\
                \textbf{RecoNet}                         &ResNet-101   & \textbf{45.54} \\
                \bottomrule
            \end{tabular}
        }}

    \end{minipage}
\end{table}

Following previous work \cite{SEGM-CVPR2018-EncNet,SEGM-CVPR2019-APCNet,SEGM-CVPR2019-CFNet,SEGM-ICCV2019-DMNet,SEGM-CVPR2019-DANet}, we use COCO-pertained model during training.
We first train our model on MS-COCO \cite{SEGM-ECCV2014-COCO} dataset for 30 epochs, where the initial learning rate is set to $0.004$.
Then the model is fine-tuned on PASCAL augmentation training set for another 80 epochs.
Finally, we fine-tune our model on original VOC12 train+val set for extra 50 epochs and the initial $lr$ is set to 1e-5.
The results in \Cref{tab:VOC_COCO} show that RecoNet-101 outperforms current state-of-the-art algorithms with the same backbone.
Moreover, RecoNet also exceeds state-of-the-art methods that use better backbone such as Xception \cite{IMGC-CVPR2017-Xception}.
By applying ResNet-152 backbone, RecoNet reaches $89.0\%$ mIoU without adding extra data.
The result is now in the $\textbf{1st}$ place of the PASCAL-VOC12 challenge\footnote{\url{http://host.robots.ox.ac.uk:8080/anonymous/PXWAVA.html}}.


\paragraph{\textbf{PASCAL-Context}.} \cite{SEGM-CVPR2014-Yang} is a densely labeled scene parsing dataset includes $59$ object and stuff classes plus one background class.
It contains $4998$ images for training and $5105$ images for testing.
Following previous works~\cite{SEGM-CVPR2019-CFNet,SEGM-CVPR2018-EncNet,SEGM-CVPR2019-APCNet}, we evaluate the dataset with background class ($60$ classes in total).
The results are shown in \Cref{tab:PContext}.
RecoNet performs better than all previous approaches that use non-local block such as CFNet and DANet, which implies that our proposed context modeling method is better than non-local block.

\paragraph{\textbf{COCO-Stuff}.} \cite{SEGM-CVPR2018-COCO-Stuff} is a challenging dataset which includes $171$ object and stuff categories.
The dataset provides $9000$ images for training and $1000$ images for testing.
The outstanding performance of RecoNet (as shown in \Cref{tab:COCO-Stuff}) illustrates that the context tensor we modeled has enough capacity to represent complicated context features.

\paragraph{\textbf{SIFT-Flow}.} \cite{SEGM-TPAMI2011-Liu} is a dataset that focuses on urban scene,
which consists of $2488$ images in training set and $500$ images for testing.
The resolution of images is $256 \times 256$ and $33$ semantic classes are annotated with pixel-level labels.
The result in \Cref{tab:SIFT-Flow} shows that the proposed RecoNet outperforms previous state-of-the-art methods.

\paragraph{\textbf{ADE20K}.} \cite{SEGM-CVPR2017-ADE20K} is a large scale scene parsing dataset which contains $25$K images annotated with $150$ semantic categories.
There are $20$K training images, $2$K validation images and $3$K test images.
The experimental results are shown in \Cref{tab:ADE20K}.
RecoNet shows better performance than non-local based methods such as CCNet \cite{SEGM-ICCV2019-CCNet}.
The superiority on result means RecoNet can collect richer context information.

\subsection{Ablation Study} 

In this section, we perform the thorough ablation experiments to investigate the effect of different components in our method and the effect of different rank number. These experiments provide more insights of our proposed method.
The experiments are conducted on PASCAL-VOC12 $validation$ set and more ablation studies can be found in supplementary material.

\noindent\textbf{Different Components.}~~~
In this part, we design several variants of our model to validate the contributions of different components. 
The experimental settings are the same with previous part.
Here we have three main components, including global pooling module (GPM) and tensor low-rank reconstruction module inducing TGM and TRM.
For fairness, we fix the tensor rank $r = 64$. The influence of each module is shown in \Cref{tab:modules}.
According to our experiment results, tensor low-rank reconstruction module contributes 9.9\% mIoU gain in network performance and the pooling module also improves mIoU by 0.6\%.
Then we use the auxiliary loss after Res-4 block. 
We finally get 81.4\% mIoU by using GPM and TGM+TRM together.
The result shows that the tensor low-rank reconstruction module dominants the entire performance. 
\begin{table}[tb!]
    \caption{Ablation study on different components. The experiments are implemented using PASCAL-VOC12 validation dataset. FT represents fine-tune on PASCAL-VOC12 original training set}
    \label{tab:modules}
    \centering
    \resizebox{.68\linewidth}{!}{\renewcommand{\arraystretch}{1.0}{
        \begin{tabular}{lccccc|c}
            \toprule
            Method     &TGM+TRM  &GPM      &Aux-loss  &MS/Flip  &FT  & mIoU \% \\  \hline \hline
            ResNet-50  &         &         &          &   &   & 68.7 \\
            ResNet-50  &$\surd$  &         &          &   &   & 78.6 \\
            ResNet-50  &$\surd$  &$\surd$  &          &   &   & 79.2 \\
            ResNet-50  &$\surd$  &$\surd$  &$\surd$   &   &   & 79.8 \\
            ResNet-101 &$\surd$  &$\surd$  &$\surd$   &   &   & 81.4 \\
            ResNet-101 &$\surd$  &$\surd$  &$\surd$   &$\surd$ &  & 82.1 \\
            ResNet-101 &$\surd$  &$\surd$  &$\surd$   &$\surd$ &$\surd$  & 82.9 \\
            \bottomrule
        \end{tabular}
    }}
\end{table}

\noindent\textbf{Tensor Rank.}~~~
Tensor rank $r$ determines the information capacity of our reconstructed attention map.
In this experiment, we use ResNet101 as the backbone.
We sample $r$ from 16 to 128 to investigate the effect of tensor rank.
An intuitive thought is that the performance would be better with the increase of $r$. However, our experiment results on \Cref{tab:modules2} illustrates that the larger $r$ does not always lead to a better performance.
Because we apply TGM+TRM on the input feature $X \in \mathbb{R}^{512 \times 64 \times 64}$, which has maximum tensor rank 64.
An enormous $r$ may increase redundancy and lead to over-fitting, which harms the network performance. 
Therefore, we choose $r=64$ in our experiments.

\begin{table}[tb!]
    \begin{minipage}{.418\linewidth}
        \caption{Ablation study on tensor rank. The results are obtained by using ResNet101 backbone and multi-scale evaluation}
        \label{tab:modules2}
        \centering
        \resizebox{4.6cm}{!}{\renewcommand{\arraystretch}{1.0}{
            \begin{tabular}{lc|c}  
                \toprule
                Method           & Tensor Rank    & mIoU \% \\  \hline \hline
                RecoNet          & 16             & 81.2 \\
                RecoNet          & 32             & 81.8 \\
                RecoNet          & 48             & 81.4 \\
                \textbf{RecoNet} & \textbf{64}    & \textbf{82.1} \\
                RecoNet          & 80             & 81.6 \\
                RecoNet          & 96             & 81.0 \\
                RecoNet          & 128            & 80.7 \\
                \bottomrule
            \end{tabular}
        }}
    \end{minipage}
    \begin{minipage}{.58\linewidth}
        \caption{Results on PASCAL-VOC12 $val$ set. RecoNet achieves the best performance with relatively small cost}
        \label{tab:VOC_comp}
        \centering
        \resizebox{5.8cm}{!}{\renewcommand{\arraystretch}{1.0}{
            \begin{tabular}{l|cc|c}  
                \toprule
                Method                                     & SS                 & MS/Flip           & FLOPs   \\  \midrule
                ResNet-101                                 & -                  & -                 & 190.6G  \\
                DeepLabV3+ \cite{SEGM-ECCV2018-DeepLabv3+} & 79.45              & 80.59             & +84.1G  \\
                PSPNet \cite{SEGM-CVPR2017-PSPNet}         & 79.20              & 80.36             & +77.5G  \\
                DANet \cite{SEGM-CVPR2019-DANet}           & 79.64              & 80.78             & +117.3G \\
                PSANet \cite{SEGM-ECCV2018-PSANet}         & 78.71              & 79.92             & +56.3G  \\
                CCNet \cite{SEGM-ICCV2019-CCNet}               & 79.51              & 80.77             & +65.3G \\
                EMANet \cite{SEGM-ICCV2019-EMANet}         & 80.09              & 81.38             & +43.1G  \\ \hline \hline
                \textbf{RecoNet}                           & \textbf{81.40}     & \textbf{82.13}    & \textbf{+41.9G} \\
                \bottomrule
            \end{tabular}
        }}
    \end{minipage}
\end{table}

\noindent\textbf{Comparison with Previous Approaches.}~~~
In this paper, we use deep-base ResNet as our backbone.
Specifically, we replace the first $7 \times 7$ convolution in ResNet with three consequent $3 \times 3$ convolutions.
This design is widely adopted in semantic segmentation and serves as the backbone network of many prior works\cite{SEGM-CVPR2017-PSPNet,SEGM-CVPR2018-EncNet,SEGM-CVPR2019-CFNet, SEGM-ICCV2019-CCNet, SEGM-ICCV2019-EMANet}.
Since the implementation details and backbones vary in different algorithms.
In order to compare our method with previous approaches in absolutely fair manner, we implemented several state-of-the-art algorithms (listed in \Cref{tab:VOC_comp}) based on our ResNet101 backbone and training setting.
The results are shown in \Cref{tab:VOC_comp}.
We compare our method with feature pyramid approaches such as PSPNet \cite{SEGM-CVPR2017-PSPNet} and DeepLabV3+ \cite{SEGM-ECCV2018-DeepLabv3+}.
The evaluation results show that our algorithm not only surpass these method in mIoU but also in FLOPs.
Also, we compare our method with non-local attention based algorithms such as DANet \cite{SEGM-CVPR2019-DANet} and PSANet \cite{SEGM-ECCV2018-PSANet}.
It is noticed that our single-scale result outperforms their multi-scale results, which implies the superiority of our method.
Additionally, we compare RecoNet with other low-cost non-local methods such as CCNet \cite{SEGM-ICCV2019-CCNet} and EMANet \cite{SEGM-ICCV2019-EMANet}, where RecoNet achieves the best performance with relatively small cost.

\subsection{Further Discussion}
We further design several experiments to show computational complexity of the proposed
method, and visualize some sub-attention maps from the reconstructed context features.

\begin{table}[tb!]
    \caption{Computational cost and GPU occupation of TGM+TRM. FLOPs (FLoating point Operations). We use tensor rank $r=64$ for evaluation}
    \label{tab:FLOPSandGPU}
    \centering
    \resizebox{.56\linewidth}{!}{\renewcommand{\arraystretch}{1.0}{
        \begin{tabular}{lccc}  
            \toprule
            Method                                   &Channel &FLOPs             &GPU Memory      \\ \hline \hline
            Non-Local \cite{ATTN-CVPR2018-Wang}      &512     &19.33G            &88.00MB         \\
            APCNet \cite{SEGM-CVPR2019-APCNet}       &512     &8.98G             &193.10MB        \\
            RCCA \cite{SEGM-ICCV2019-CCNet}          &512     &5.37G             &41.33MB         \\
            $A^2$Net \cite{ATTN-NIPS2018-A2-Nets}    &512     &4.30G             &25.00MB         \\
            AFNB \cite{SEGM-ICCV2019-AsymNL}         &512     &2.62G             &25.93MB         \\
            LatentGNN \cite{ATTN-ICML2019-LatentGNN} &512     &2.58G             &44.69MB         \\ 
            EMAUnit \cite{SEGM-ICCV2019-EMANet}      &512     &2.42G             &24.12MB         \\ \hline \hline
            \textbf{TGM+TRM}                         &512     &\textbf{0.0215G}  &\textbf{8.31MB} \\ \bottomrule
        \end{tabular}
    }}
\end{table}

\noindent\textbf{Computational Complexity Analysis.}~~~
Our proposed method is based on the low-rank tensors, thus having large advantage on computational consumption.
Recalling that non-local block has computational complexity of $\mathcal{O}(CH^2W^2)$.
On the TGM stage, we generates a series of learnable vectors using $1 \times 1$ convolutions.
The computational complexity is $\mathcal{O}(C^2 + H^2 + W^2)$ while on the TRM stage, we reconstruct the high-rank tensor from these vectors and the complexity is $\mathcal{O}(CHW)$ for each rank-1 tensor.
Since $CHW >> C^2 > H^2 = W^2$, the total complexity is $\mathcal{O}(rCHW)$, which is much smaller than non-local block. Here $r$ is the tensor rank.
\Cref{tab:FLOPSandGPU} shows the FLOPs and GPU occupation of TGM+TRM.
From the table we can see that the cost of TGM+TRM is neglegible compared with other non-local based methods.
Our proposed method has about \textbf{900} times less FLOPs and more than \textbf{100} times less FLOPs compared with non-local block and other non-local-based methods, such as $A^2$Net \cite{ATTN-NIPS2018-A2-Nets} and LatentGNN \cite{ATTN-ICML2019-LatentGNN}.
Besides of these methods, we calculate the FLOPs and GPU occupation of RCCA, AFNB and EMAUnit, which is core component of CCNet \cite{SEGM-ICCV2019-CCNet}, AsymmetricNL\cite{SEGM-ICCV2019-AsymNL} and EMANet\cite{SEGM-ICCV2019-EMANet}. It can be found that TGM+TRM has the lowest computational overhead. 

\begin{figure*}[tb!] 
    \centering 
    \includegraphics[width=.96\linewidth]{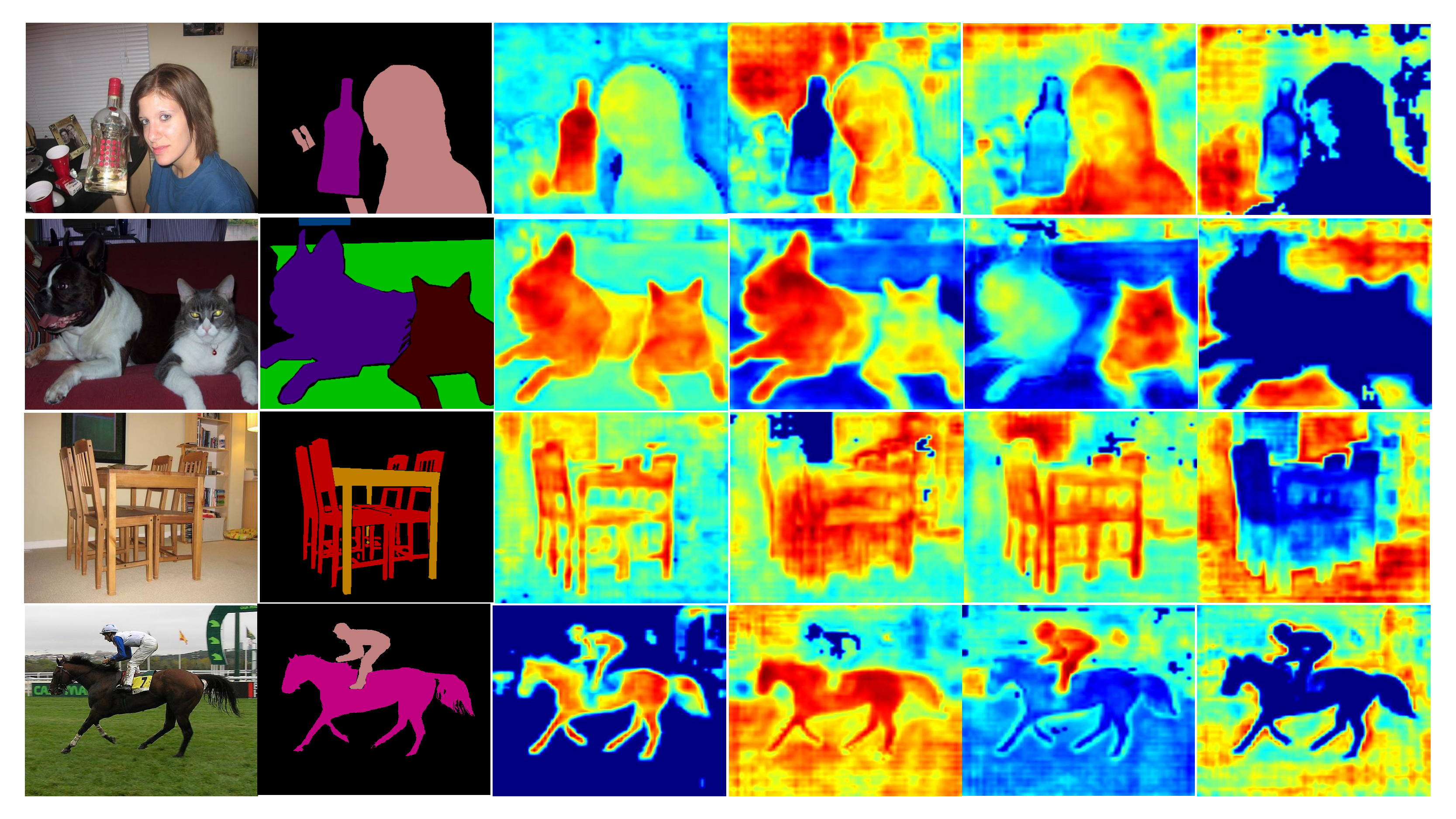} 
    \caption{Visualization of sub-attention map. From left to right are Image, Ground Truth, $\vec{A}_i \cdot \Vec{X}$, $\vec{A}_j \cdot \Vec{X}$, $\vec{A}_k \cdot \Vec{X}$, and $\vec{A}_l \cdot \Vec{X}$. It can be found that sub-attention maps mainly focus on the different parts of image. }
    \label{fig:Tensor Visualization}
\end{figure*}

\noindent\textbf{Visualization.}~~~
In our proposed method, context features are constructed by the linear combination of sub-attention maps, \ie, $\vec{A}_i \cdot \Vec{X}$.
Therefore, we visualize their heat maps to check the part of features they activate.
We randomly select four sub-attention maps $\vec{A}_i \cdot \Vec{X}$, $\vec{A}_j \cdot \Vec{X}$, $\vec{A}_k \cdot \Vec{X}$, $\vec{A}_l \cdot \Vec{X}$, as shown in \Cref{fig:Tensor Visualization}.
We can see that different sub-attention maps activate different parts of the image.
For instance, for the last case, the four attention maps focus on the foreground, the horse, the person, and the background, respectively,
which implies that the low-rank attention captures the context fragments and RecoNet can catch long-range dependencies.


\section{Conclusion}
\label{sec:conclu}
In this paper, we propose a tensor low-rank reconstruction for context features prediction, which overcomes the feature compression problem that occurred in previous works.
We collect high-rank context information by using low-rank context fragments that generated by our proposed tensor generation module.
Then we use CP reconstruction to build up high-rank context features.
We embed the fine-grained context features into our proposed RecoNet.
The state-of-the-arts performance on different datasets and the superiority on computational consumption show the success of our context collection method.

%

\clearpage
\section{Appendix}
\subsection{More Experimental Results}


We conduct experiments on Cityscapes dataset \cite{SEGM-CVPR2016-Cityscapes}, which is a famous scene segementation dataset that includes 19 semantic classes. It provides 2975/500/1525 images for training, validation and testing. Since the training setting of Cityscapes is very distinct to the implementation details that presented in main paper, we put the results in supplementary materials. 

The input images are cropped into 512 $\times$ 1024 before input. The batch size we use is 8. Initially, the learning rate $lr = 0.01$. SGD optimizer with momentum = 0.9 and weight decay = 0.0005 is applied for training. The evaluation metrics and data augmentation strategies we use are the same as main paper.
\begin{table}[htbp]
 	\centering
	\caption{Results on Cityscapes $test$ set}
    \begin{tabular}{lc|c}  
        \toprule
        Method                               &Backbone    & mIoU \\           \hline \hline
        PSANet \cite{SEGM-ECCV2018-PSANet}     &ResNet-101  & 80.1 \\
        CFNet \cite{SEGM-CVPR2019-CFNet}         &ResNet-101  & 79.6 \\
        AsymmetricNL \cite{SEGM-ICCV2019-AsymNL}   &ResNet-101 & 81.3 \\
        CCNet \cite{SEGM-ICCV2019-CCNet}   &ResNet-101 & 81.4 \\
        DANet \cite{SEGM-CVPR2019-DANet}     &ResNet-101 & 81.5 \\
        ACFNet \cite{SEGM-ICCV2019-ACFNet}         &ResNet-101 & 81.8 \\
        
         \hline \hline
        \textbf{RecoNet}                     &ResNet-101 &\textbf{82.3} \\
        \bottomrule
    \end{tabular}
    \label{tab:cityscape}
\end{table}
For the evaluation on $val/test$ set, we train 40K/100K iterations on $train/train+val$ set respectively. The testing results are shown on \Cref{tab:cityscape}, which collects current state-of-the-art attention based methods. RecoNet get better performance than these approaches. The online hard example mining (OHEM) strategy is not used in our implementation since it is time consuming. The result is avaliable on the website.\footnote{\url{https://www.cityscapes-dataset.com/anonymous-results/?id=7c7bfabc1026a9fd07b348bfd311c56a57ba0369969f3bd9fd9f036ce49a2934}}

In order to validate the consistency of RecoNet, we conduct additional ablation experiments on Cityscapes dataset. The tensor rank is set to $r = 64$ for ablation. In \Cref{tab:additioanl}, it can be found that TGM+TRM contributes 5.8 \% mIoU improvement (73.1\% to 78.9\%), which dominates the other modules. The experimental results show that RecoNet is consistent on different datasets.

\begin{table}[tb!]
    \caption{Ablation study on different components. The experiments are implemented using Cityscapes validation set}
    \label{tab:additioanl}
    \centering
    \resizebox{.68\linewidth}{!}{\renewcommand{\arraystretch}{1.0}{
        \begin{tabular}{lcccc|c}
            \toprule
            Method     &TGM+TRM  &GPM      &Aux-loss  &MS/Flip    & mIoU \% \\  \hline \hline
            ResNet-50  &         &         &          &$\surd$     & 73.1 \\
            ResNet-50  &$\surd$  &         &          &$\surd$     & 78.9 \\
            ResNet-50  &$\surd$  &$\surd$  &          &$\surd$     & 79.4 \\
            ResNet-50  &$\surd$  &$\surd$  &$\surd$   &$\surd$     & 79.8 \\
            ResNet-101 &$\surd$  &$\surd$  &$\surd$   &     & 80.5 \\
            ResNet-101 &$\surd$  &$\surd$  &$\surd$   &$\surd$  & 81.6 \\
            \bottomrule
        \end{tabular}
    }}
\end{table}

\subsection{More Visualization}
\begin{figure}[tb!] 
    \centering 
    \includegraphics[width=0.718\linewidth]{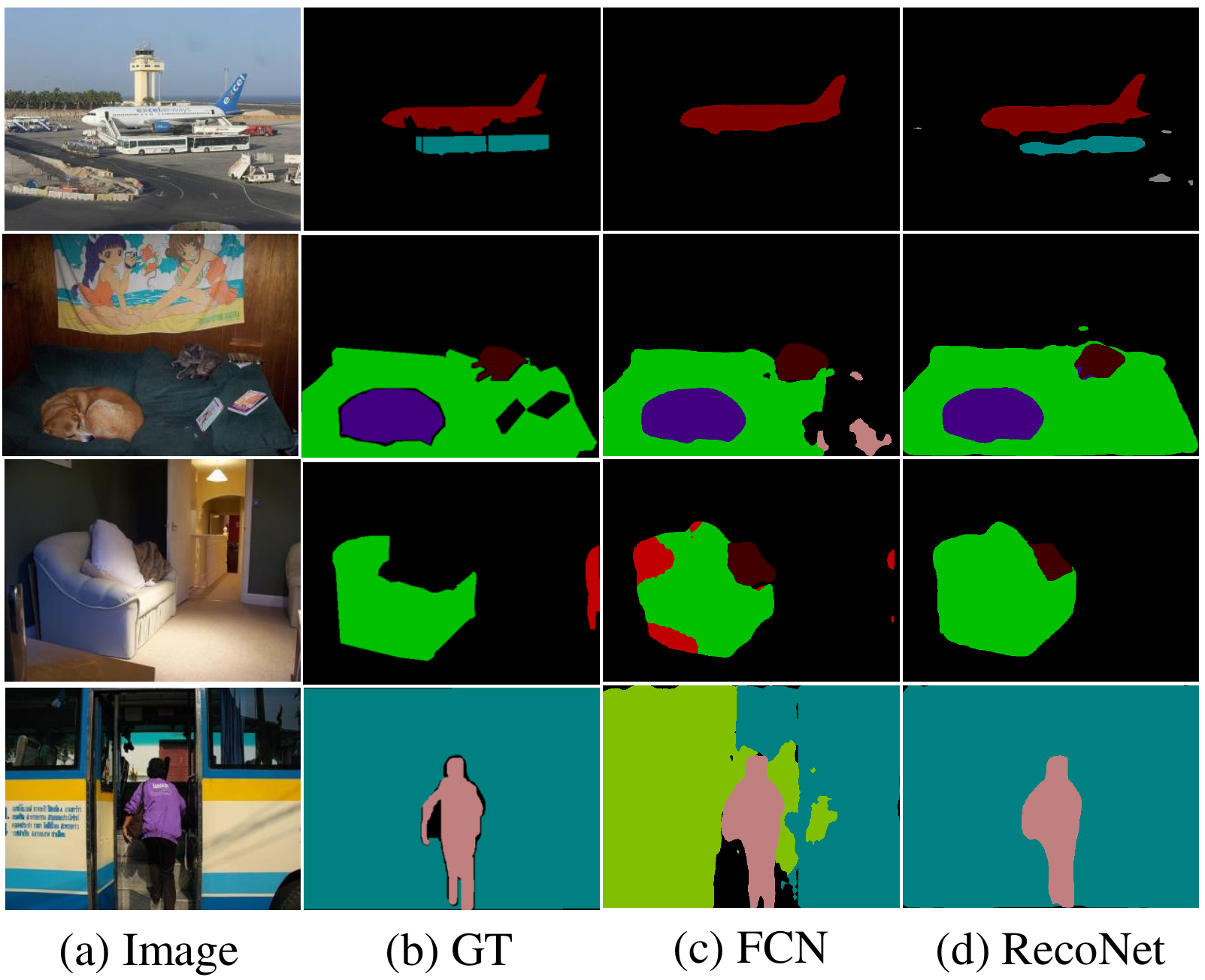} 
    \caption{Qualitative results on PASCAL-VOC12 $validation$ dataset.} 
    \label{fig:Result Visualization}
\end{figure}

\Cref{fig:Result Visualization} shows some results of RecoNet-101 on PACAL-VOC12 $validation$ dataset.
The figure shows that RecoNet has a better qualitative result, especially in the boundary, which also demonstrates its effectiveness of context modeling.

%
%
\bibliographystyle{splncs04}
\bibliography{Top-sim,CV,SEGM,DL,SPEED,TNSR}

\end{document}